\documentclass[letterpaper, 10 pt, conference]{ieeeconf}  

\IEEEoverridecommandlockouts                              

\usepackage{amssymb}
\usepackage{subcaption}
\usepackage[fleqn]{amsmath}
\usepackage{graphicx}
\usepackage{multirow}
\usepackage{floatrow}
\usepackage{amsfonts}
\usepackage{hyperref}
\usepackage{svg}
\usepackage[ruled]{algorithm2e}

\title{\LARGE \bf
Reinforcement Learning-based Virtual Fixtures for Teleoperation of Hydraulic Construction Machine*
}

\author{ Hyung Joo Lee$^{1}$ and Sigrid Brell-Cokcan$^{1}$
\thanks{*This work has been supported by the North Rhine-Westphalia Ministry of Economic Affairs, Innovation, Digitalisation and Energy of the Federal Republic of Germany under the research intent 5G.NAMICO: Networked, Adaptive Mining and Construction.}
\thanks{$^{1}$The authors are with Chair of Individualized Production (IP), RWTH Aachen University, Campus-Boulevard 30, 52074 Aachen, Germany.
        {\tt\small {lee, brell-cokcan}@ip.rwth-aachen.de}}%
}

\begin{document}

\maketitle
\thispagestyle{empty}
\pagestyle{empty}

\begin{abstract}
The utilization of teleoperation is a crucial aspect of the construction industry, as it enables operators to control machines safely from a distance. However, remote operation of these machines at a joint level using individual joysticks necessitates extensive training for operators to achieve proficiency due to their multiple degrees of freedom. Additionally, verifying the machine's resulting motion is only possible after execution, making optimal control challenging. In addressing this issue, this study proposes a reinforcement learning-based approach to optimize task performance. The control policy acquired through learning is used to provide instructions on efficiently controlling and coordinating multiple joints. To evaluate the effectiveness of the proposed framework, a user study is conducted with a Brokk 170 construction machine by assessing its performance in a typical construction task involving inserting a chisel into a borehole. The effectiveness of the proposed framework is evaluated by comparing the performance of participants in the presence and absence of virtual fixtures. This study's results demonstrate the proposed framework's potential in enhancing the teleoperation process in the construction industry.

\end{abstract}


\section{INTRODUCTION}

In contrast to the production industry, the construction industry presents a unique challenge to automation due to its dynamic and multifaceted nature \cite{lee1}. Construction sites are constantly changing through various stages of development, where each stage possesses different dangerous conditions for human workers \cite{brell}. To mitigate these risks, teleoperation has become a vital component of construction machinery in the current construction industry \cite{chen}. However, the complex nature of these machines, which often have multiple degrees of freedom and require individual levers for remote control on a joint level, requires significant operator training. Even experienced operators may require months of training to coordinate multiple joints to achieve the desired end-effector or tool motion. This can result in decreased productivity and reduced local accuracy and work efficiency.

While intelligent robotic systems with advanced control algorithms have been developed to address the challenges posed by the unstructured nature of construction sites and provide numerous benefits associated with automation \cite{joo, hutter2}, the autonomy of these robots may be limited in highly unstructured environments. This is because incomplete and inaccurate information about unknown objects or unexpected situations can significantly impact the decision-making process of the robots, compromising their ability to operate autonomously.
\begin{figure}[!t] 
\centering
\includegraphics[scale=0.15]{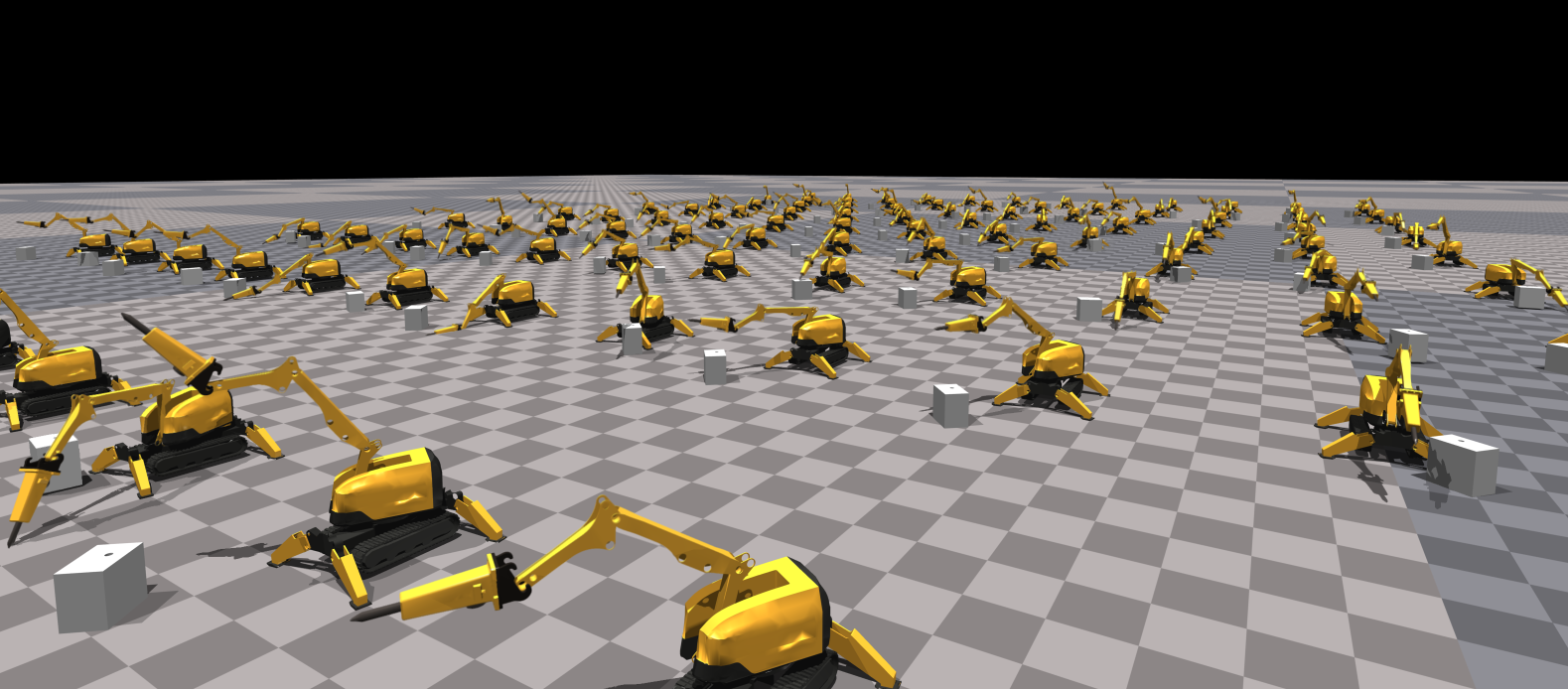}
\caption{Simulation environment with 128 agents being progressed in parallel to train a control policy for the hammer insertion task. }
\label{isaacgym}
\end{figure}
To overcome the limitations of autonomous robots in unstructured environments, research is focused on enhancing operator capabilities through teleoperation systems that integrate automation techniques such as virtual fixtures \cite{rosenberg}. Virtual fixtures guide a remote manipulator or robot to perform a desired task and can be classified as forbidden-region virtual fixtures or guidance virtual fixtures \cite{Bettini}. These systems reduce operator workload, increase task efficiency, and improve system performance in complex tasks \cite{rosenberg2}. However, the complex system dynamics of hydraulic construction machines present a significant challenge in this area of research \cite{feng}. Analyzing and modeling the system's behavior is often challenging due to its significant nonlinear system dynamics.

To address these challenges, we suggest employing a data-oriented method that leverages data gathered while the machine operates, enabling the capture of its unique attributes. Our method does not depend on a detailed analytical model of the system, time-consuming parameter tuning, or costly hardware adjustments. The machine's data-driven model is seamlessly integrated into a reinforcement learning (RL) framework. This approach empowers RL agents to discover efficient approaches for accomplishing tasks autonomously, without human intervention. The policies generated by these agents can then assist human operators by guiding virtual fixtures. The paper is organized as follows: First, we propose an RL framework to solve the hammer insertion task. Also, the data-driven actuator model, combined with the dynamic simulator to generate a close-to-reality actuator motion, is also proposed. Section 4 introduces the virtual fixtures method that utilizes the trained control policy from the proposed RL framework. In Section 5, we highlight the capabilities of the proposed system through the hammer insertion task and a user study. Subsequently, we draw essential conclusions and outline future works.

\section{Related Work}
Developing an autonomous robotic system that can effectively operate in unstructured environments is a formidable challenge, owing to the complexity and limitations inherent in such environments. As a result, the interest in teleoperation support as a viable alternative has continued to grow. Among the promising methods of teleoperation support is the virtual fixture-based assistance system, first introduced by Rosenberg \cite{rosenberg}. This system enhances situational awareness and mitigates the cognitive load of teleoperation by overlaying supplementary information on sensory data. Since the introduction of Rosenberg's collaborative control concept in teleoperation, numerous other approaches have been proposed \cite{ryden,21}.

While virtual fixtures have been shown to enhance system performance in previous studies, limited research has focused on applying virtual fixtures to construction machines. The dynamic nature of construction sites, with frequently changing task conditions and objectives, presents a significant challenge. A pre-defined set of virtual fixtures can quickly become impractical and restrictive. Furthermore, precise information about the system dynamics is often necessary to effectively guide operators using virtual fixtures. However, modeling the system dynamics of construction machines is often challenging due to the nonlinearity of their hydraulic circuits, requiring significant effort and resources that are frequently impractical.

Applying data-driven methods in nonlinear construction machines has collected considerable interest among researchers. Notably, researchers have employed neural networks to capture the underlying dynamics of excavators using a supervised learning approach \cite{exc1}, and have trained control policies for grading tasks through a data-driven actuator model in simulation using a reinforcement learning framework \cite{egli}. Despite the advancements established in these studies, human operators do not use the knowledge generated about optimizing excavator control. Given construction sites' unstructured and dynamic nature, it can be challenging to implement autonomous approaches for all construction sites. Thus, transferring the knowledge from the autonomous control policies to the operator can be advantageous. This transfer would allow operators to benefit from the optimized movement learned through the autonomous approach and apply this knowledge to subsequent teleoperation tasks.

This study aims to expand upon the conventional data-driven reinforcement learning (RL) approach for training control policies. We propose a framework where RL agents initially learn effective task-completion strategies without human intervention. Subsequently, these learned policies are utilized to teach humans by providing visualized optimized control inputs, including joystick motions, with guidance virtual fixtures. The primary objective of this approach is to introduce a new method for supporting operators in the teleoperation of construction machines by leveraging the latest advancements in data-driven control methods. Specifically, the operator specifies the overarching objective by designing the reward function, and the RL agents generate efficient skills that the operator can learn via the guidance of virtual fixtures. We demonstrate the effectiveness of our proposed framework by conducting hardware experiments on Brokk 170, a hydraulic construction machine in the context of a hammer insertion task, and present the results of our user study.

\section{Method}

\begin{figure}[!t] 
\centering
\includegraphics[scale=0.35]{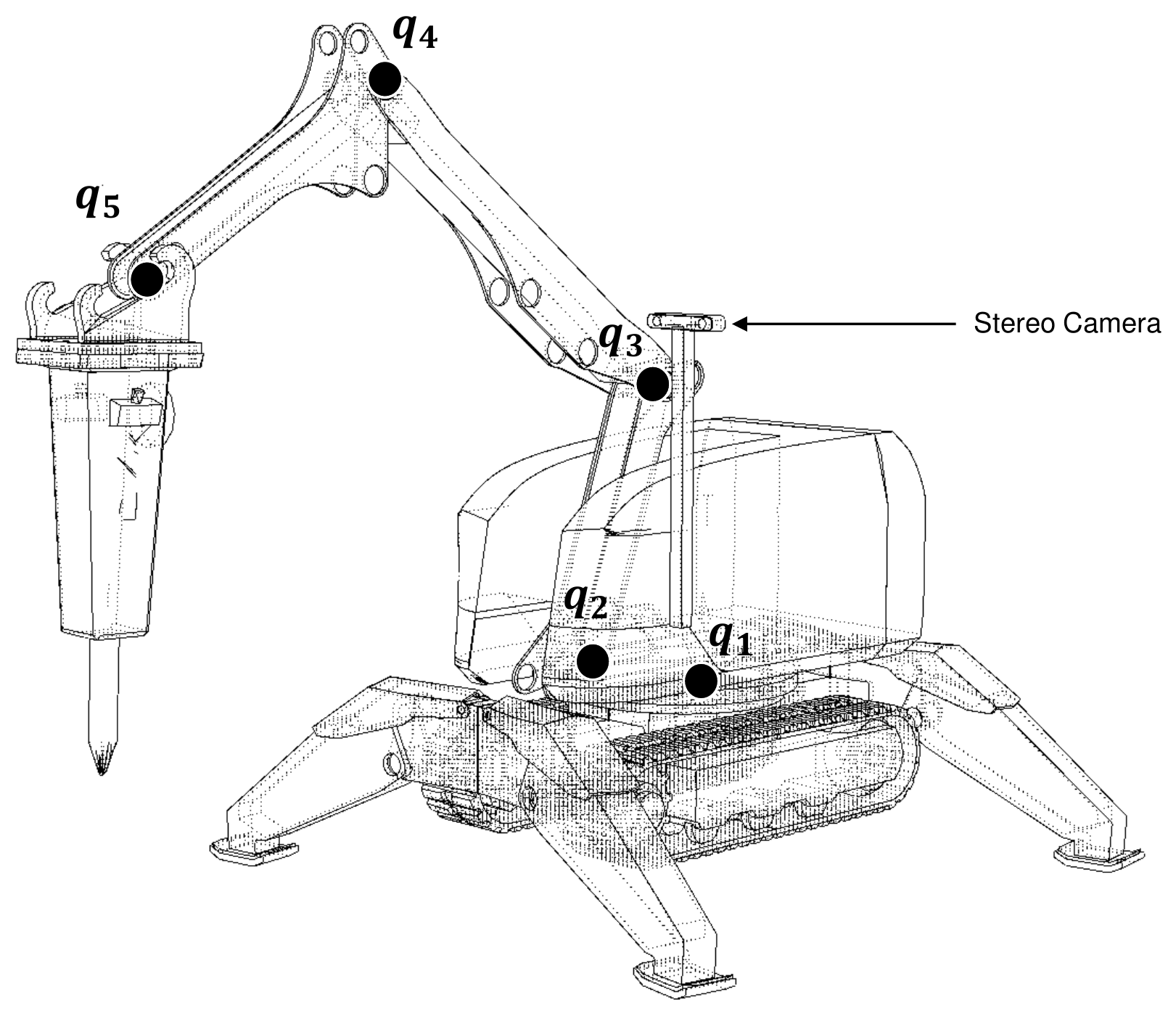}
\caption{Joint configuration of Brokk 170 used in this work.}
\label {joint_config}
\end{figure}

To facilitate the transfer of knowledge from machines to humans, it is necessary to first train the construction machine to effectively tackle the given task, considering that the task conditions and the construction site undergo constant changes throughout the project. Data-centric methodologies, such as reinforcement learning (RL), have gained significant traction in addressing this challenge, presenting a versatile learning framework for machines. RL involves the accumulation of data through iterative experimentation and automatic fine-tuning of the control policy to optimize the reward function that represents the task \cite{sutto}. This process can be fully automated, enabling end-to-end optimization of the control policy, from capturing sensor readings to generating precise low-level control signals, thereby creating a highly adaptable framework for identifying suitable skills to resolve the task at hand. However, RL typically necessitates extensive interaction time with the system to acquire complex skills, often spanning hours or even days of real-time execution \cite{levine}. Furthermore, during the training phase, the machine may exhibit abrupt and unpredictable behavior, giving rise to safety concerns, particularly in scenarios involving heavy-duty machinery.

The proposed methodology aims to tackle the challenge of effectively training construction machines to handle intricate tasks by leveraging advanced physics simulation technology. This involves training the machines in a simulated environment and transferring the acquired skills to human operators in real-world scenarios. This approach reduces the risk of unpredictable machine behavior at the beginning of the training and lessens the need for extensive training data. However, its effectiveness depends on successfully bridging the reality gap between the simulated and real-world systems. To address this challenge, this work utilizes the concept of the data-driven actuator model \cite{Hwangbo} to address the discrepancies in system dynamics between the simulation and real-world environments.

\begin{figure}[!t] 
\centering
\includegraphics[scale=0.2]{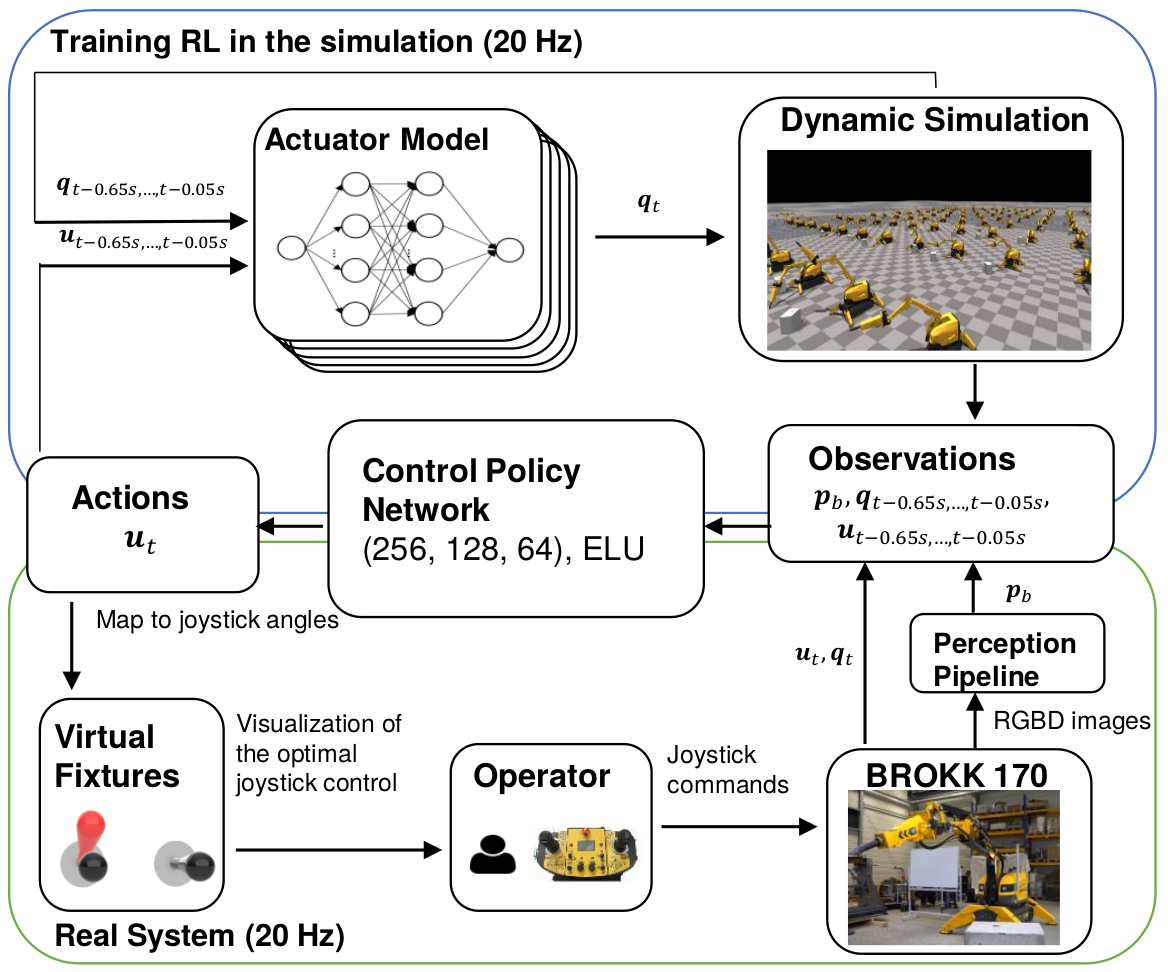}
\caption{The approach starts by training a control policy in a dynamic simulation with the actuator model to minimize the gap between reality and simulation. Here, the actuator model accepts $u_{t{-0.65s}, ..., t-0.05s}$ and $q_{t{-0.65s}, ..., t-0.05s}$ as inputs and generates $q_t$. This output is subsequently utilized within the dynamic simulation to reproduce a realistic actuator motion. The policy, trained using this actuator model, is then implemented on the physical machine. The resulting output is converted to joystick angles for virtual fixtures to the operator. By following the control policy's computed outputs as a guideline, the operator learns to perform the task optimally.}
\label {overview}
\end{figure}

The proposed approach involves steps illustrated in Figure \ref{overview}. Firstly, the actuation is modeled using measurements obtained during the operation of the construction machine. Secondly, a control policy is trained in the simulation for the given task, taking into account the underlying dynamics of the machine in conjunction with the actuator model. Thirdly, a perception pipeline is implemented to estimate the pose of the borehole on the test concrete block. Finally, the trained control policy is applied to the real machine, and the output of the control policy is transformed into virtual guidance fixtures to train human operators in efficiently accomplishing the task.

\subsection{Data-Driven Actuator Model}
This study explores the relationship between control input and system state change in hydraulic machines by utilizing a multi-layer perceptron (MLP) neural network. The MLP has two hidden layers, each with 32 units, and is equipped with a rectified linear unit (ReLu) activation function and a linear output layer. To gather training data that effectively characterizes the behavior of the actuator at varying directions and velocities, we generated PWM signals in a sine waveform. The frequency and amplitudes of the sine wave were randomly adjusted. The resulting PWM signals were applied until the associated joint reached its maximum or minimum position, at which point the joint was returned to its original home configuration. Training data is collected at 20 Hz covering the entire actuation range. The duration of the data collection process is 1.0$h$ for each joint. To minimize the risk of collision between the machine and the environment, the training data is collected separately for each joint since the coupling dynamics between joints were found to be insignificant. However, in the case of substantial coupling dynamics, the data must be collected by simultaneously actuating multiple joints to consider the impact of hydraulic coupling. Moreover, the actuator model should incorporate a unified and comprehensive network capable of representing multiple actuators alongside their inherent dynamics, including hydraulic coupling effects.

\begin{figure}[!t] 
\centering
\includegraphics[scale=0.14]{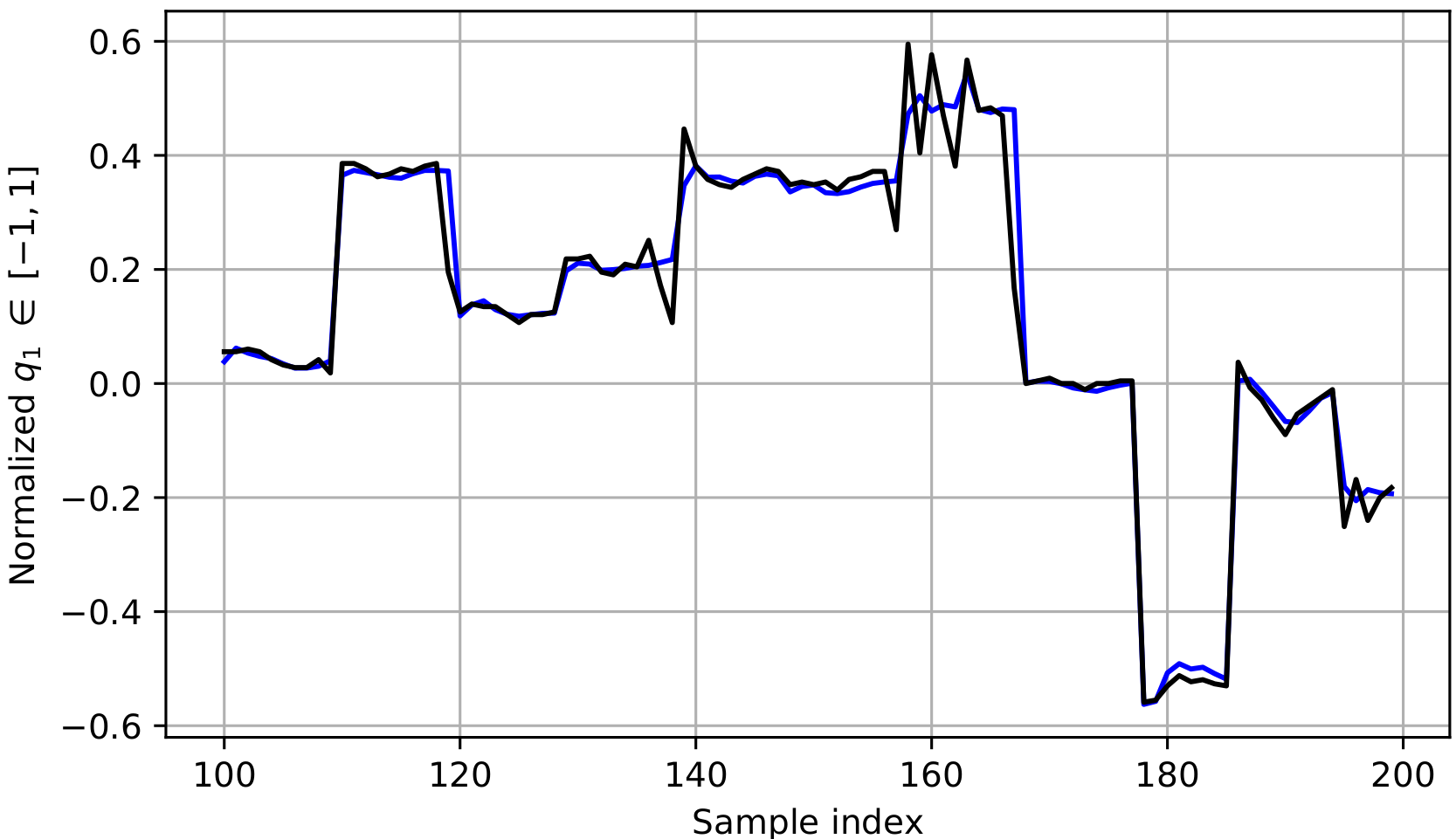}
\caption{Comparison between the collected (black) and predicted $q_1$ (blue).}
\label {prediction}
\end{figure}

Here, the input to the actuator model is designed to encompass the entire system delay, which is experimentally determined to be 550$ms$. Thus, the history of the control input and system state from the preceding 650$ms$ are incorporated into the inputs. The output is the corresponding system state change. To facilitate the training process, all input and output data are normalized. The model is trained using the mean squared error (MSE) loss function and the Adam optimizer. The loss function converges after roughly 1$h$ of training time, resulting in 5$h$ of total training time for all 5 joints. Fig. \ref{prediction} illustrates the results of the trained actuator model, where the predicted $q_1$ is directly compared with the collected $q_1$. The results demonstrate that the trained actuator model can predict $q_1$ at different speeds with direction changes.

\subsection{Learning to solve the hammer insertion task}
First, we set up the desired task in a simulated environment. Our ultimate objective is to obtain a control policy that can generate control inputs for each of the joints $q_{1...5}$ to insert a hammer into a borehole. It is important to note that joint $q_2$ cannot be actuated simultaneously with other joints due to the manufacturer-defined design of the machine. Therefore, the control policy must learn how to independently control $q_2$ and $q_{1,3,4,5}$ to insert an 8 cm diameter chisel into a 10 cm diameter borehole. Furthermore, the actuator model trained in Sec. III-A is integrated into the simulation to allow the control policy to effectively manage the machine's input delays and hydraulic properties and coordinate the joints optimally. In this task, the borehole's position is detected via a stereo camera (see Sec. III-C). We assume the borehole's orientation is fixed since the concrete block used in this task cannot be tilted during the experiments. Fig. \ref{overview} provides an overview of the proposed system. 

\subsubsection{States and Observation }
At each time step $t$, the agent receives an observation $s_t$ that includes information about the current state of the environment, such as the borehole position $p^b$ and the same information that the actuator network requires, as shown in Table 1. Providing the agent with a history of past measurements is vital to enable optimal coordination of different joint delays. Additionally, the agent receives the position of the borehole (i.e., the target position). The agent's actions consist of control inputs for each joint provided to the actuator model in the simulation during training or the actual machine during deployment (see Table II). We normalize all observations and actions to accelerate training with an approximate mean and standard deviation. In practical settings, the inferred position of the borehole from stereo cameras is subject to noise. To account for this, we incorporate uniformly sampled white noise to $p^b$, with a maximum amplitude of $5\%$. Furthermore, to address potential inaccuracies in the concrete block model, we introduce random variations in the collision geometry of the block by increasing its size by up to $10\%$. To improve the robustness of our policy and to account for errors in the actuator model, we also introduce noise to the other components $q$ and $u$. Specifically, we randomly scale the values of $q$ and $u$ by a maximum of $10\%$, and this scaling factor remains constant throughout a given episode.

\begin{table}
\begin{tabular}{l|l|lllllllll}
\hline
Observation                              &    Description &    Dimensions        \\ \hline  \hline 
$p_{b}$              & Borehole position & 3   \\
$q_{t{-0.65s}, ..., t-0.05s}$  & Joint positions            & 65    \\
$u_{t{-0.65s}, ..., t-0.05s}$  & Control inputs             & 65    \\ \hline
\multicolumn{3}{l}{}\\
\hline Actions                              &    Description &    Dimensions        \\ \hline  \hline 
$u_{t}$              & Control inputs & 5   \\ \hline
\end{tabular}
\caption{ Observations and actions of the agent }
\end{table}

\subsubsection{Disjoint control}
To ensure effective control of the hydraulic machine, teaching the agent the appropriate actions is essential. Since the machine cannot move all joints simultaneously, $q_2$ must be controlled separately as it is used primarily to adjust the distance to the target object. To facilitate this, we introduce action masks, namely $[0,1,0,0,0]$ and $[1,0,1,1,1]$, which restrict the agent's actions to prevent simultaneous movements of all joints. We also encourage the agent to adjust the distance to the target object at the beginning of the operation, following the expert behavior of human operators. Accordingly, an action mask $[0,1,0,0,0]$ is applied to the computed control input $u_t$ so that only $q_2$ is actuated until the target object is within a good reachable working area. Since $q_1$ is solely responsible for rotating the base, the remaining joints, $q_{3,4,5}$, are used for the manipulation task. Empirical analysis has shown that the manipulator has good manipulability over the target object when the distance $d_t$ between $q_3$ and the target object is within $1.7 \leq d_t \leq 2.4$ meters. Hence, the agent stops actuating $q_2$ and utilizes $q_{1,3,4,5}$ when $1.7 \leq d_t \leq 2.4$ or the height of the end-effector $d_h$ is below 0.2 meters to prevent collisions with the ground.

\begin{figure*}[!t] 
\centering
\includegraphics[scale=0.45]{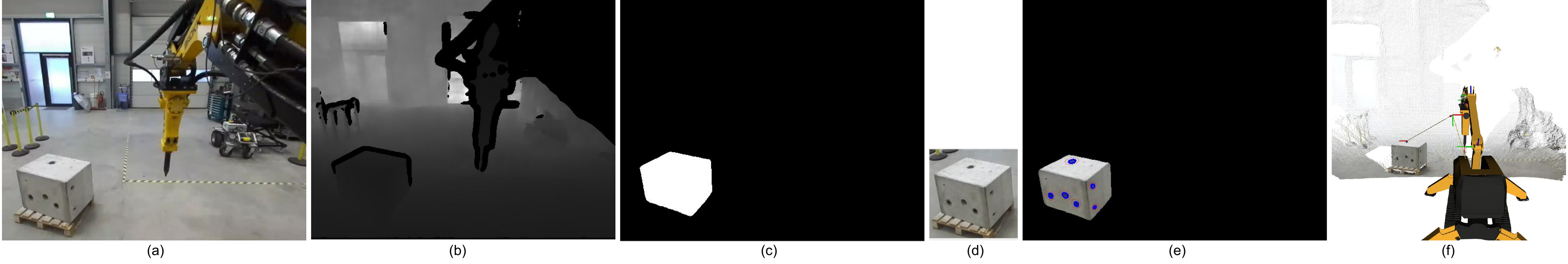}
\caption{Overview of how the borehole position is estimated. First, the concrete block is segmented from the retrieved RGB (a) images using the trained U-net (c). By using the mask of the segmented concrete block, the area around the concrete block is cropped (d). Ellipse-like shapes are detected from the cropped RGB images to ensure that only the target concrete boreholes are considered the final borehole position. The borehole positions are estimated by finding the corresponding depth value from the identified ellipse center point (e). By comparing the normals, only the borehole of the interest is considered (f).}
\label {visual_pipeline}
\end{figure*}

\begin{figure}[!t] 
\centering
\includegraphics[scale=0.33]{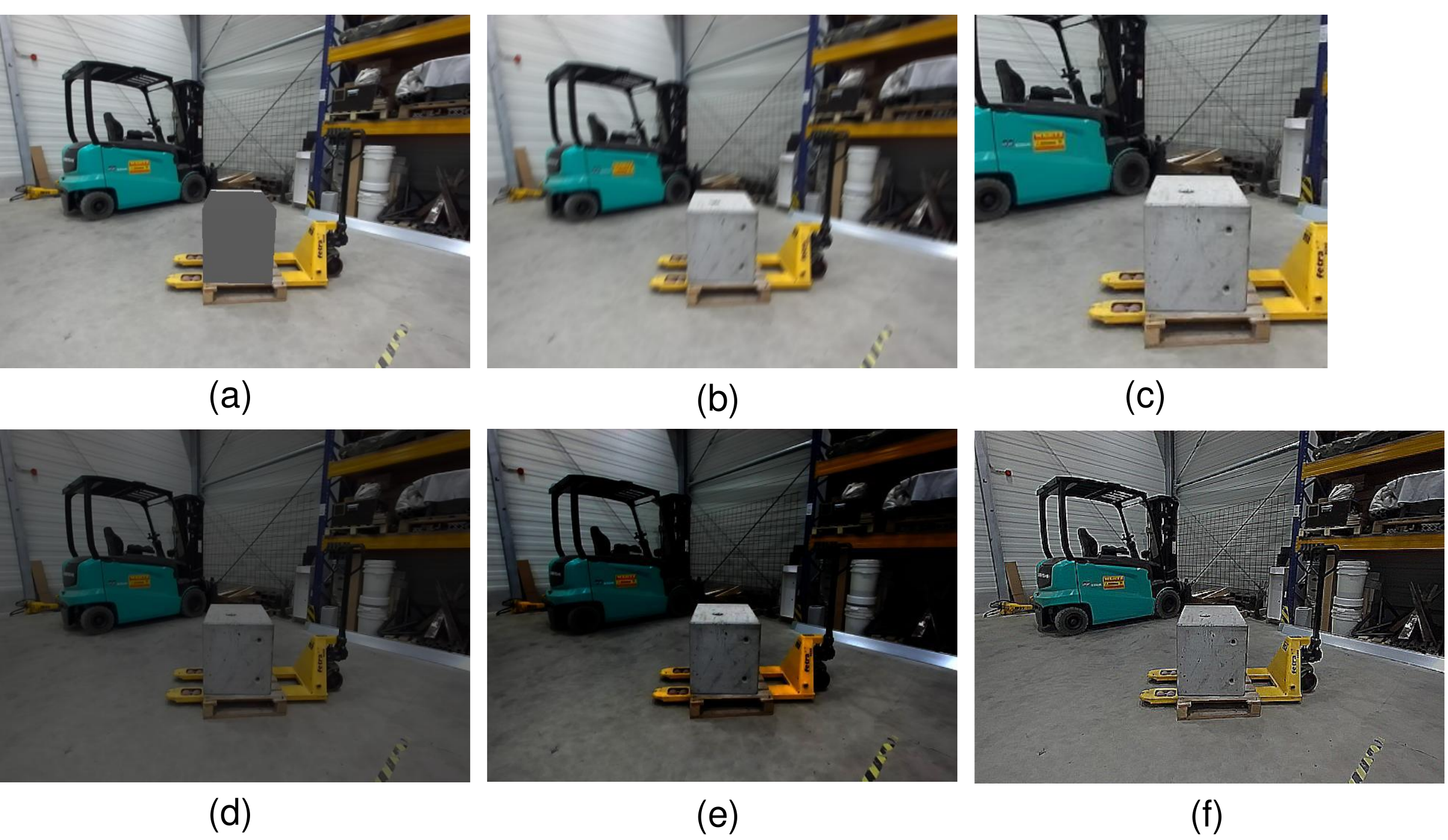}
\caption{Labeled data generated by the pipeline (a). Each labeled data is augmented with blurred (b), cropped (c), light contrast (d), gamma (e), and noise (f) effects. Here, only the data (a) is collected by a hand-carrying RGBD camera.}
\label {bbb}
\end{figure}

\subsubsection{Rewards}

A reward function must be implemented to guide the learning process to achieve a desired behavior in the control policy. This function should impose penalties on actions that result in undesired behaviors, such as collisions with concrete blocks. The reward functions are designed such that the agent utilizes joint $q_2$ to position the manipulator in a location that facilitates object manipulation while utilizing joints $q_{1,3,4,5}$ to insert the hammer into the borehole. This approach avoids unnecessary collisions with the environment and promotes smooth commands by penalizing significant changes in control actions:

\begin{equation}
    r_t= 
\begin{cases}
    r_t^m, & \text{if } 1.7 \geq d_t \text{ or } d_t \geq 2.4   \\
    & \text{             or } d_{h} \geq 0.2 \\
    r_t^d + r_t^o + r_t^i - r_t^c - r_t^a,              & \text{otherwise}
\end{cases}
\end{equation}
where
\begin{align*} 
    r_t^m = 
    \begin{cases}
     0.25/(1+d_t-2.4), & \text{if } d_t \geq 2.4 \\
     0.25/(1+1.7-d_t),              & d_t \leq 1.7
\end{cases}
\end{align*}
\vspace{-0.5cm}
\begin{align*} 
    r_t^d = 1/(1+\lVert p_t^b - p_t^{ee} \rVert_2),
\end{align*}
\vspace{-0.5cm}
\begin{align*} 
    r_t^o = 1/(1+|1.57 - \theta_{t}^{ee}|),
\end{align*}
\vspace{-0.5cm}
\begin{align*} 
    r_t^i= 
\begin{cases}
    5(r_t^d+r_t^o), & \text{if } r_t^d, r_t^o \geq 0.9  \text{ and } p_t^{ee_z} \leq 0.7 \\
    0,              & \text{otherwise}
\end{cases}
\end{align*}
\vspace{-0.4cm}
\begin{align*} 
    r_t^c = \begin{cases}
    0.01f_t^{ee}, &       \text{if } r_t^d \leq 0.7 \text{ and } \lVert f_t^{ee} \rVert_2 \geq 0\\
    0,              & \text{otherwise}
\end{cases}
\end{align*}
\vspace{-0.5cm}
\begin{align*} 
    r_t^a = 0.01\lVert u_t - u_{t-1} \rVert_1,
\end{align*}

At each time step $t$, this study's total reward $r_t$ comprises six distinct terms, outlined in equation (1). The first term, denoted as $r_t^m$, encourages the agent to move the manipulator to an optimal position for effective object manipulation. The following two terms, $r_t^d$ and $r_t^o$, encourage the agent to approach the borehole while aligning the hammer perpendicular to the surface. The estimated borehole depth $p_t^{b_z}$ is reduced by 0.1 $m$ to encourage the agent to move inside the borehole. The desired hammer orientation for successful insertion is assumed to be $\phi^{ee},\theta^{ee},\psi^{ee} = (0, 1.57, 0)$ since the borehole orientation is always parallel to the $xy$-plane. A successful hammer insertion is rewarded by the term $r_t^i$, while collisions are penalized by the term $r_t^c$, but only if the agent is far from the borehole. Contact is detected by measuring the hammer's force value $f_t^{ee}$. To ensure smooth command execution during operation, the term $r_t^a$ penalizes large changes in control actions. This is important because the agent's commands are transmitted to a human operator in the final stage (Section III-D) to assist them, and sudden, significant changes in commands are found to be difficult for the operator to follow.

\subsubsection{Training Procedure}
To achieve effective training of the agent, a high sampling rate is crucial for learning the specified tasks. In this study, we utilized the IsaacGym simulator \cite{isaacgym}, a simulation environment designed to enable policy learning with a high sampling rate by parallelizing physics on a single GPU. Following each episode, we randomly set the initial arm configuration and borehole position to enable the trained agent to perform the task from various positions. We terminated an episode prematurely if the machine was at its joint limits or if the maximum episode length was reached. Additionally, after the random reset of the arm configuration and borehole, there were instances where the starting arm configuration intersected with the concrete block. In such cases, we terminated the episode immediately with zero rewards.

To train the control policy for the hammer insertion task, the Proximal Policy Optimization (PPO) algorithm is used \cite{ppo}. Specifically, the PPO implementation provided by Makoviichuk et al. \cite{rlgames} is used, which supports GPU-accelerated training with parallel environments using Isaac GYM. Two multilayer perceptron (MLP) neural networks approximate the value function and the policy, each with Exponential Linear Unit (ELU) activation in the hidden layers and a linear output layer. The networks consist of 3 hidden layers, with 256 units in the first layer, 128 units in the second layer, and 64 units in the last layer. The control policy is trained at 20 Hz on a computer with an Intel Core i7-9850H CPU, 16GB of RAM, and an NVIDIA Quadro RTX 3000 GPU. The training process takes approximately 3 hours to converge, during which 128 agents are used in parallel to accelerate the training process. 

\subsection{Pose Estimation of the Borehole}
In this study, we utilize an RGBD camera to estimate the borehole position observed by the trained agent (see Table I), focusing on the boreholes on the target objects to ensure that the agent only considers those as target points. The workflow is illustrated in Fig. \ref{visual_pipeline}. Firstly, RGB and depth images are captured using a Zed2i camera. Next, a U-net segmentation network, which utilizes the LabelFusion framework \cite{labelfusion}, is employed to segment the concrete block from the RGB image \cite{unet}. To ensure robustness in different lighting and noise conditions, we also incorporate an additional data augmentation pipeline to generate the different training data sets, as shown in Fig. \ref{bbb}. After the concrete block is segmented, the corresponding mask in the RGB image is used to restrict the interest area by cropping the image. We employ an ellipse detection framework from \cite{opencv} on the cropped RGB image to detect the boreholes. By performing the ellipse detection only on the cropped image, we ensure that only the boreholes from the interest area (i.e., where the target concrete block is detected) are considered the target position. The center point of the detected ellipses is estimated, and the corresponding depth values are retrieved from the depth image. We limit the target borehole to the one on the $xy$-plane and filter out the other ellipses by comparing the normal values. Finally, the center point of the final ellipse is considered the target position by the agent and is also visualized by the operator.

\section{Virtual Fixtures}
The aim of this work is to generate virtual fixtures that guide a human operator in a hammer insertion task using a trained control policy. The policy takes as input the borehole position and the joint positions and control inputs from the previous 650 milliseconds, as indicated in Table I, and provides the next control inputs, which are scaled to the range of [-1, 1]. Here, the computed next control input is converted into the desired lever angle $\alpha_t^d$ as follows:
\begin{equation}
    \hspace{2.5cm}\alpha_t^d = \alpha_{max}u_t 
\end{equation}
where the $\alpha_{max}$ represents the maximum lever angle from the real lever. The left lever is responsible for the $q_1$ and $q_4$, whereas the right lever controls $q_{2,3,5}$. As multiple joints need to be controlled with each lever, (2) is extended to represent the desired lever orientation in two different orientations:

\begin{align}
    &(l\_\phi^d, l\_\theta^d, l\_\psi^d) = (0, \alpha_t^{d_4}, \alpha_t^{d_1})\\
    &(r\_\phi^d, r\_\theta^d, r\_\psi^d)=
    \begin{cases}
    (0, \alpha_t^{d_2}, 0), & \text{if } u_t^{d_2} \neq 0 \\
    (0, \alpha_t^{d_3}, \alpha_t^{d_5})             & \text{otherwise}
    \end{cases}
\end{align}

\begin{figure}[!t] 
\centering
\includegraphics[scale=0.085]{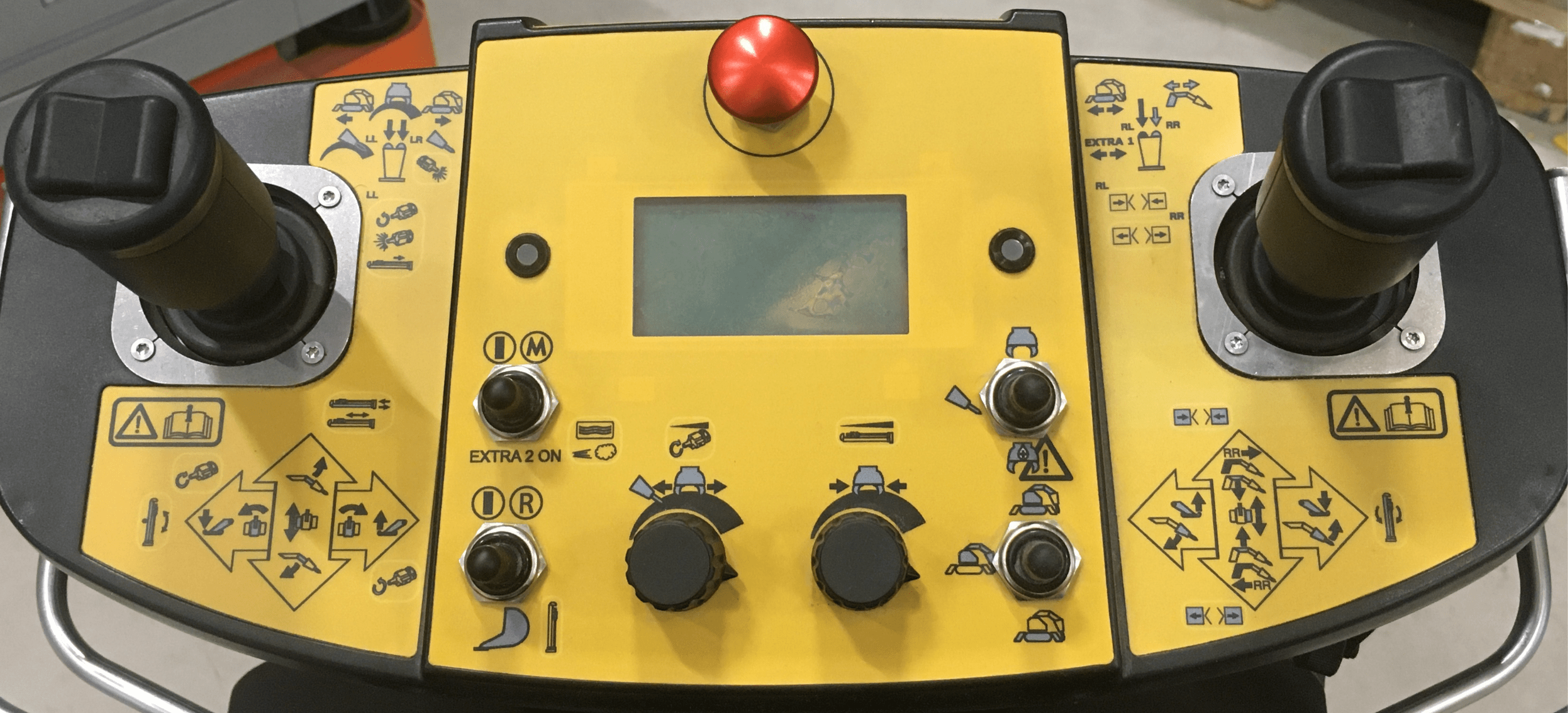}
\caption{Control device for Brokk 170. The left lever controls $q_{1,4}$ and the right lever $q_{2,3,5}$. When the right button on the right lever is activated, the right lever controls $q_2$ and otherwise $q_{3,5}$.}
\label {joystick}
\end{figure}

\begin{figure}[!t] 
\centering
\includegraphics[scale=0.25]{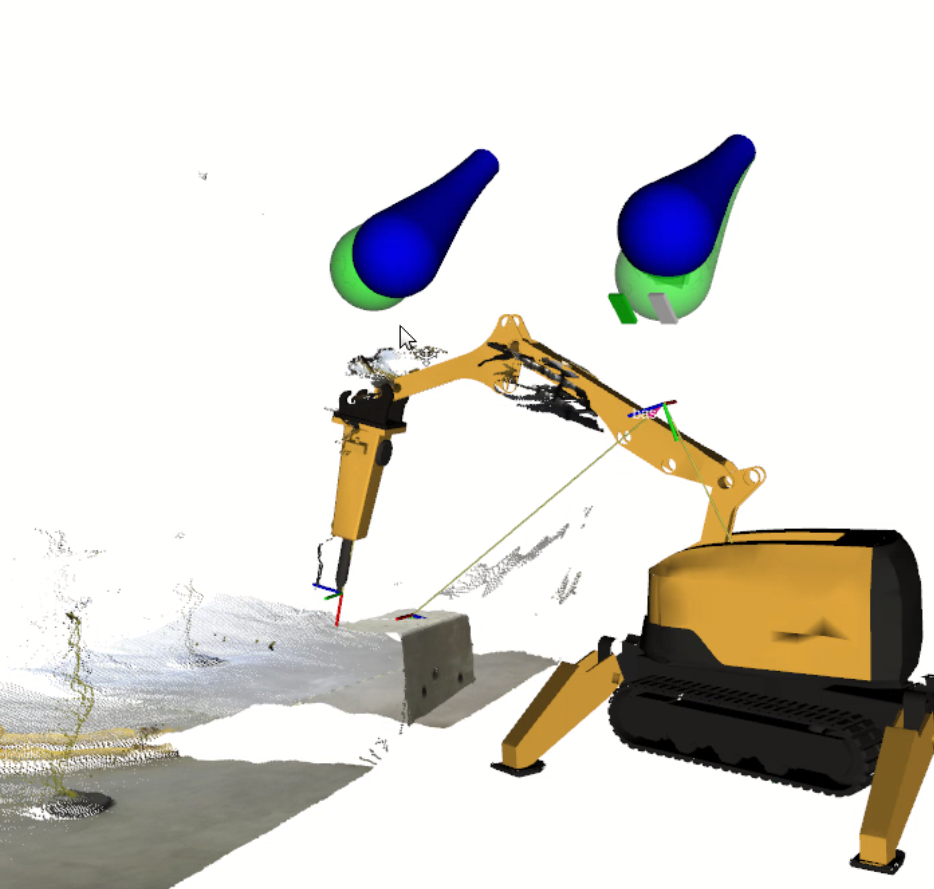}
\caption{The virtual fixtures with the 3D point cloud are provided to the operator. The computed joystick maneuvers from the trained control policy are shown in green, whereas the operator's maneuvers are visualized in blue.}
\label{rviz}
\end{figure}

Notably, the right lever controls three joints, where $q_2$ is separately controlled from other joints (see Fig. \ref{joystick}). The buttons on the right lever allow the control of $q_2$ to be switched between the right lever and $q_{3,5}$. To facilitate the lever control process, the visualization of the desired lever control is overlaid with the actual lever control, enabling the operator to compare both and more easily mimic the desired lever control (see Fig. \ref{rviz}). This approach is designed to alleviate the burden on human operators and reduce the risk of errors by leveraging the knowledge from the obtained control policy to teach an optimized lever control.

\section{Experimental Evaluation}

\begin{figure}[!t] 
\centering
\includegraphics[scale=0.35]{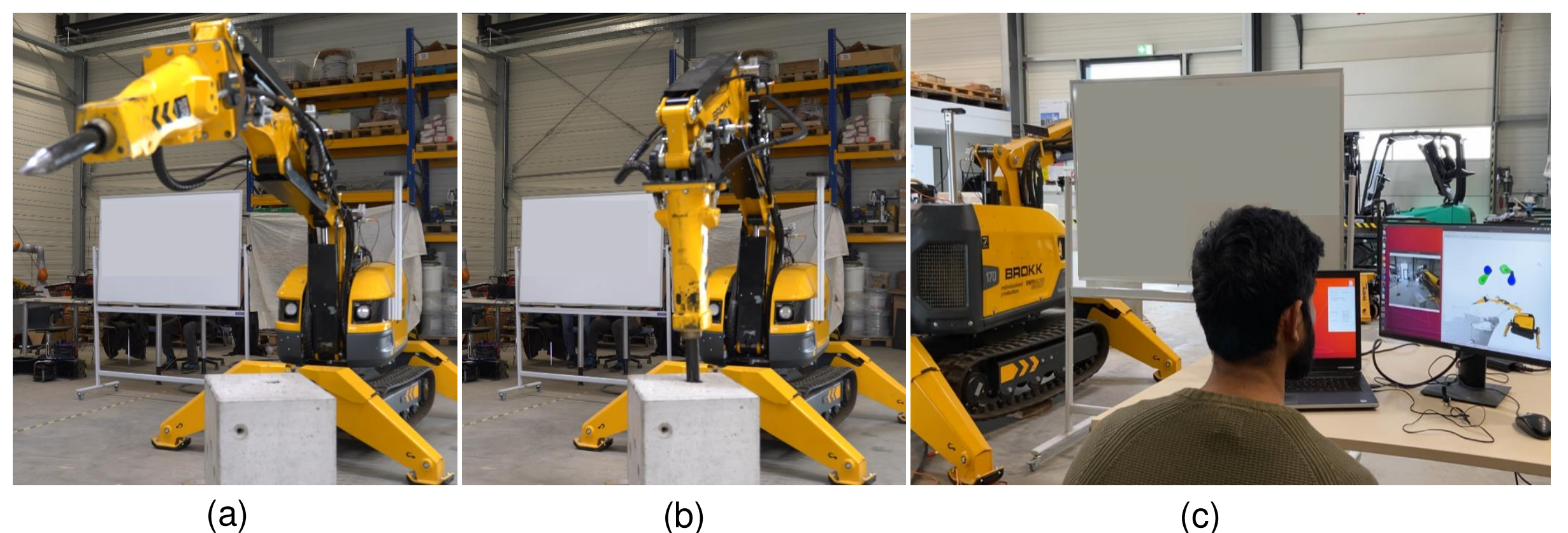}
\caption{Experimental setup with an arbitrary start position (a). Using the provided virtual fixtures, the operator inserts the hammer into the borehole (b). The direct sight into the workspace is limited to emulate the scenario where the operator remotely works in a safe control room (c).}
\label {subject}
\end{figure}

The evaluation of the proposed approach involves a user study to demonstrate the impact of virtual fixtures on performance. The study includes 5 participants who have no experience with the teleoperation of Brokk 170. The study consists of two experiments with and without virtual fixtures, each requiring the insertion of a hammer with a diameter of 8cm into a 10cm diameter borehole located at different positions on a concrete block. To perform the task, the operator uses 2D RGB images and 3D point cloud data provided at 3 Hz (see Fig. \ref{subject}). The borehole is detected only once at the beginning of each experiment due to its detection dependency on the manipulator's configuration. The control policy, which computes the optimal joystick maneuver at 20Hz, utilizes the detected borehole position.

In the first experiment, the participants are instructed to perform hammer insertion using virtual fixtures. The detected borehole position is sent to the trained control policy along with the joint positions and control inputs from the last 650 ms. The control policy then calculates the corresponding control input towards the detected borehole position, converted to the corresponding joystick angles. The green joystick visualizes the optimal joystick maneuver from the trained control policy, while the blue joysticks represent the current maneuver. To guide the operator when to use $q_2$, the color of the buttons on the right joystick is changed to green. The participants perform the same task in the second experiment without the virtual fixtures. Before starting the experiment, each participant is provided with a brief introduction to the teleoperation of Brokk 170 and given 10 minutes of practice time to become familiar with it. To evaluate the system's performance qualitatively, the participants fill out NASA Task-Load Index (TLX) questionnaires \cite{tlx}. Additionally, the end-effector motion is recorded during the experiment by applying forward kinematics with the joint angle values.

\begin{figure}[!t] 
\centering
\includegraphics[scale=0.44]{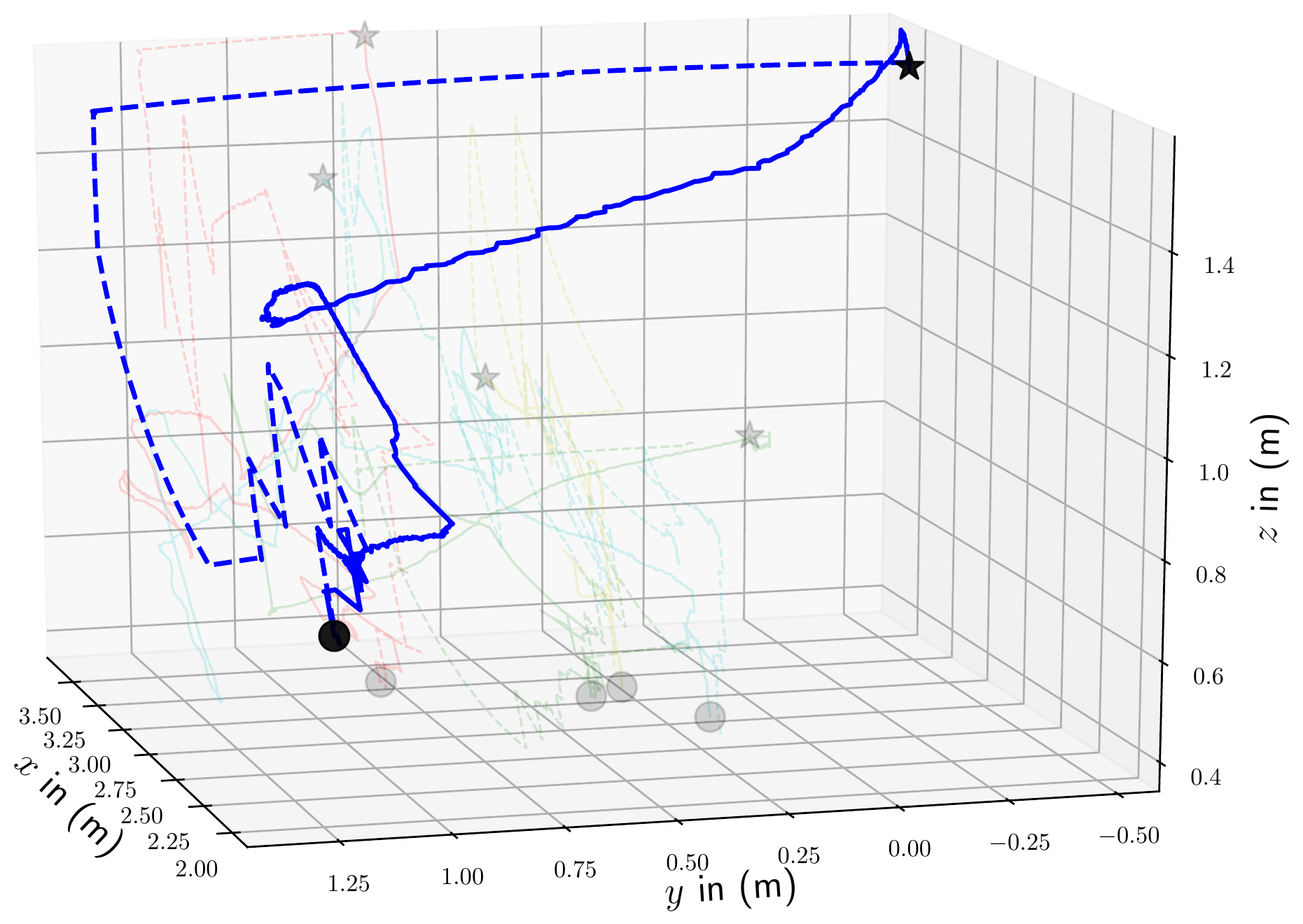}
\caption{Recorded end-effector path during the user study. The starting and borehole positions are changed for each participant and visualized with stars and spheres, respectively. Solid: virtual fixtures. Dotted: without virtual fixtures.}
\label {userstudy_trajectory}
\end{figure}

\begin{figure}[!t] 
\centering
\includegraphics[scale=0.6]{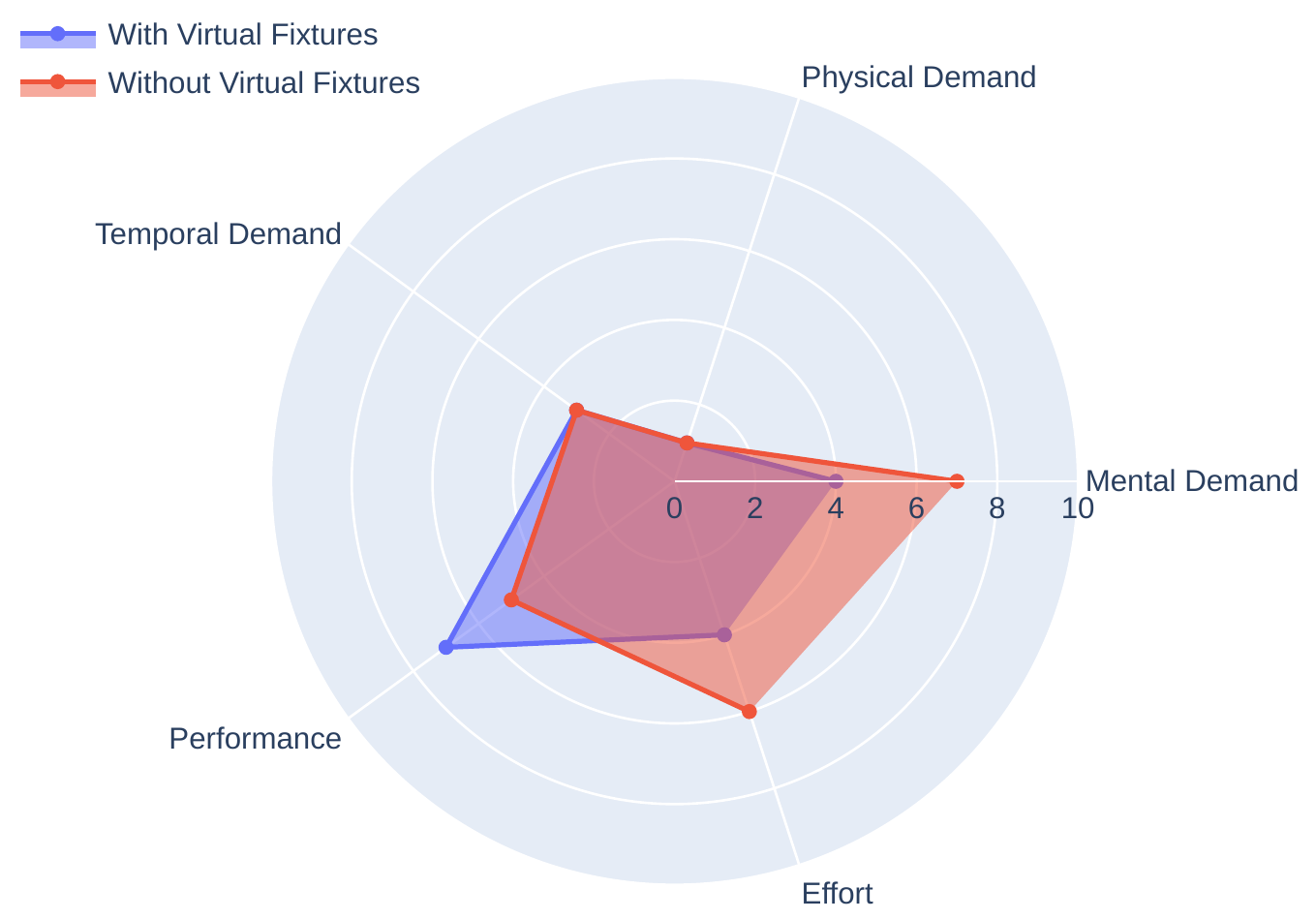}
\caption{Results from the NASA-TLX questionnaires in radar graph.}
\label{userstudy}
\end{figure}

Figure \ref{userstudy_trajectory} illustrates the recorded trajectory of the end-effector during the hammer insertion task, wherein the outcome of a single participant is emphasized to enhance visual clarity. The solid curve corresponds to the virtual fixtures-based experiment, while the dotted curve represents the experiment without virtual fixtures. The star and sphere icons depict the start and end positions of the task, respectively, which vary across participants. The blue trace is used to facilitate the comparison for a particular participant. Moreover, Figure \ref{userstudy} presents the mean scores of the NASA Task-Load Index (TLX) questionnaires completed by the five participants in both virtual fixture-based and non-virtual fixture-based experiments.

The user study results demonstrate a notable disparity in the teleoperation of the machine. As depicted in Figure \ref{userstudy_trajectory}, the participants struggled to control multiple joints simultaneously and preferred to control them individually. This difficulty was particularly evident in the borehole area, where the participants had difficulty determining the correct joint angles to insert the hammer. However, when presented with visualized virtual fixtures derived from the trained control policy, the participants tended to control multiple joints simultaneously, resulting in a more efficient path to the borehole. The computed $q_2$ motion ensured optimal distance to the borehole, reducing the number of movements required for hammer insertion. This difference is also evident in Figure \ref{userstudy}. The findings from the questionnaires indicated that the utilization of virtual fixtures resulted in a reduction in mental demands and required effort for the participants. This reduction occurred as the fixtures provided clear guidelines for optimal joystick utilization during teleoperation. However, it should be noted that additional exploration and evaluation involving a larger number of participants are needed to assess the impact on mental burden comprehensively.

\section{CONCLUSIONS}
This paper presents an RL-based framework that can efficiently learn the hammer insertion task in a simulation environment. To minimize the simulation-to-reality gap, we use a data-driven actuator model during training that captures the machine-specific nonlinearities in the relationship between control inputs and system state changes. The learned control policy then guides human operators in optimal joint control for task performance. In teleoperation, the experience of human operators heavily influences task performance, as they can only verify the machine's resulting motion after execution. To address this issue, we propose an assistance system that provides guidelines for optimal joystick maneuvers and joint combinations through virtual fixtures. By doing so, safe teleoperation can be ensured, and errors during the task can also be minimized. We demonstrate our framework's effectiveness in inserting an 8cm diameter chisel into a 10cm diameter borehole. Our method is evaluated through a user study, which shows promising results in reducing the mental burden and effort required to perform the task. Furthermore, the resulting end-effector path shows that virtual fixtures allow the task to be performed with fewer motions, as operators tend to use multiple joints with the aid of virtual fixtures simultaneously.

In future work, we plan to enhance the visual feedback provided in our framework. Currently, the visual information is streamed at a rate of 3 Hz, which heavily burdens the WLAN network due to the streaming of 3D point clouds and 2D RGB images. To address this issue, we are exploring the integration of 5G technology at our test construction site to enable more efficient visual feedback during teleoperation \cite{5g}. Additionally, we aim to extend our framework by integrating a mobile platform for capturing visual information. In the current setup, occlusions frequently occur due to the fixed camera position. By incorporating a mobile platform capable of dynamically navigating around the workspace to obtain the best camera view based on the relative pose between the construction machine and the target object, we intend to improve further the visual feedback provided to the operator. Furthermore, our future plans include extending the proposed approach to encompass various tasks. In the current stage, the agent within the RL framework is focused on learning a specific task, namely the insertion of a hammer into a borehole. However, by instructing the agent in the general approach to maneuver the manipulator towards the desired goal pose, considering the underlying dynamics and the coordination of different joints, the proposed system can be utilized more effectively for multiple tasks.


\addtolength{\textheight}{-12cm}   





\end{document}